\documentclass{vgtc}                          

\graphicspath{{figures/}{pictures/}{images/}{./}} 

\usepackage{times}                     
 
\usepackage{tabu}                      
\usepackage{booktabs}                  
\usepackage{lipsum}                    
\usepackage{mwe}                       
\usepackage{amssymb}
\usepackage{mathptmx}                  
\usepackage{booktabs}
\usepackage{multirow}
\usepackage{amsmath}
\usepackage[inkscapelatex=false]{svg}

\onlineid{1433}

\vgtccategory{Research}

\usepackage[table,xcdraw]{xcolor}
\definecolor{NavyBlue}{RGB}{0,0,128} 
\vgtcinsertpkg

\usepackage{hyperref}

\title{S-Avatar : Diffusion-Guided Gaussian Head Avatars from a Single Image
}

\author{Hail Song\thanks{e-mail: hail96@kaist.ac.kr}\\
\scriptsize KAIST UVR Lab%
\and Seokhwan Yang\thanks{e-mail: 
ysshwan147@kaist.ac.kr}\\
\scriptsize KAIST UVR Lab%
\and Jiwon Yang\thanks{e-mail: 
yjw514@kaist.ac.kr}\\
\scriptsize KAIST UVR Lab%
\and Woojin Cho\thanks{e-mail: woojin.cho@kaist.ac.kr}\\
\scriptsize KAIST UVR Lab%
\and Woontack Woo\thanks{e-mail: wwoo@kaist.ac.kr}\\ %
     \parbox{1.4in}{\scriptsize \centering KAIST UVR Lab \\ KAIST KI-ITC ARRC}}

\teaser{
  \centering
  \includegraphics[width=\linewidth]{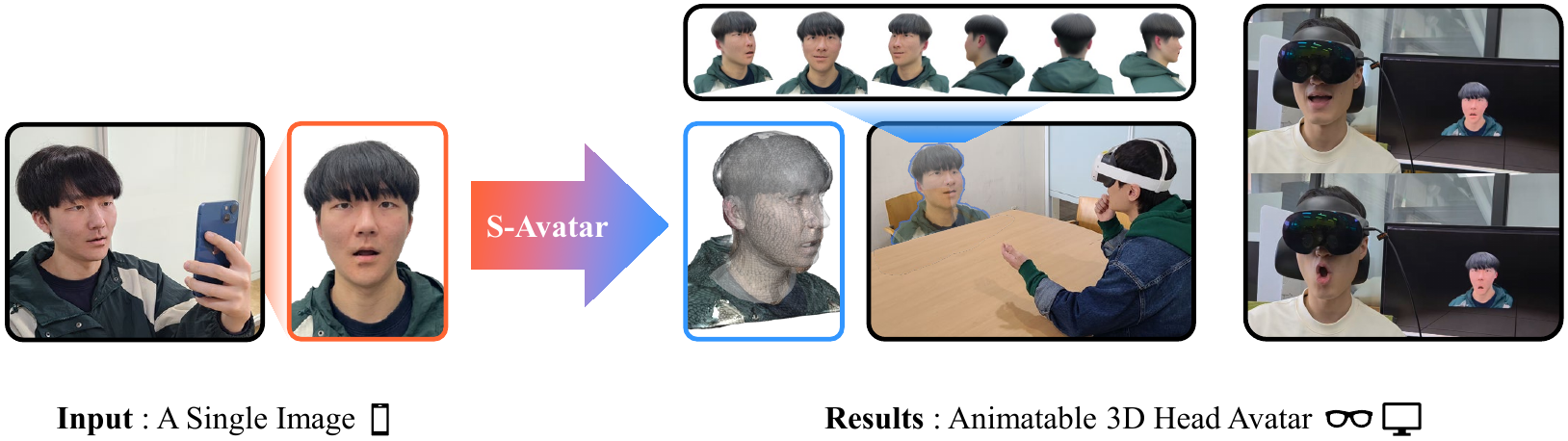}
\caption{We introduce \textbf{S-Avatar}, a novel method for generating 3D human head avatars from a single image input. Utilizing a diffusion-based Gaussian splat generation module and a fitting/binding strategy for dynamic 3D Gaussian Splatting with expression control, our approach enables one-shot head avatar generation. This method produces photorealistic avatars that can be rendered from novel viewpoints and controlled with diverse facial expressions. The avatars generated by S-Avatar are applicable to Virtual and Augmented Reality (VR/AR) applications.}

  \label{fig:teaser}
}

\abstract{
We propose \textbf{S-Avatar}, a novel method for generating photorealistic 3D head avatars from a single image using a diffusion-guided 3D model generation module and strategies for animating 3D Gaussian Splatting (3DGS). While single-image head avatar reconstruction is crucial for lifelike Virtual Reality (VR) applications, existing approaches often struggle to preserve 3D consistency under unseen viewpoints. 
S-Avatar addresses this limitation through a three-stage pipeline. First, a high-resolution 3DGS is synthesized directly from a single image using a diffusion-based Gaussian splat generation module. Next, the parametric head model FLAME is aligned with the generated 3DGS by optimizing its parameters and spatial transformations. Finally, to adapt the 3DGS to FLAME variations, we construct a binding template that encodes the spatial relationship between the initial splats and FLAME. The dynamic 3D head avatar can then be rendered in real time by deforming the 3DGS with the binding template. 
By combining diffusion-guided canonical 3DGS generation with FLAME-based control, our method achieves efficient and accurate reconstruction with enhanced 3D consistency. Evaluations on public datasets demonstrate that S-Avatar outperforms state-of-the-art methods in novel-view and expression generation, achieving superior realism and consistency. Consequently, our approach represents a significant advance in accessible avatar creation, applicable to a wide range of VR/AR applications. The project page is available at \url{https://github.com/hailsong/savatar}.
}


\begin{document}
\maketitle

\section{Introduction}
In Virtual Reality (VR) and Augmented Reality (AR), modeling a human avatar that closely resembles the user's real appearance is a critical component. 
In particular, photorealistic head avatars are essential for effective communication in virtual environments~\cite{ma2021pixel}.
Realism in VR/AR avatars hinges on controllability and multi-view rendering, both of which are vital for accurate 3D identity representation.
Furthermore, there is a growing demand for single-image avatar reconstruction methods, which offer a convenient and widely accessible solution without the need for complex capturing equipment.

Numerous studies aiming to reconstruct human avatars reflecting the user's appearance are based on the parametric model. The 3-Dimensional Morphable Models (3DMMs)~\cite{3dmm} enable the detailed modeling and realistic simulation of human faces by blending different facial attributes and expressions from a set of predefined models. Complementing the capabilities of 3DMMs in facial representation, the FLAME model~\cite{flame} extends these principles to include the entire head by efficiently and effectively capturing various facial shapes, poses, and expressions.
Studies on video-based 3D head avatar reconstruction track facial movements and depth information from sequential frames to generate highly detailed models~\cite{garrido2016reconstructionvideoface, tewari2019fmlvideoface, tewari2021learning}. Meanwhile, considering the convenience of data acquisition, single-image avatar reconstruction has also been actively studied, primarily by combining parametric facial models such as FLAME or 3DMM with neural networks to estimate 3D shapes~\cite{feng2021deca, zielonka2022mica, danvevcek2022emoca, deng2019accurate, sanyal2019learning}. However, these parametric model-based reconstruction methods have limitations in expressiveness, making it difficult to accurately capture fine details such as hair and intricate skin textures.

For detailed avatar representation, recent studies have proposed approaches that integrate advanced neural rendering techniques with parametric head models.
Thanks to the remarkable performance of Neural Radiance Fields (NeRF)~\cite{mildenhall2021nerf} and 3D Gaussian Splatting (3DGS)~\cite{3dgaussiansplatting} in constructing 3D representations from multi-view observations, several methods have adopted NeRF~\cite{xu2023latentavatar, imavatar} and 3DGS~\cite{chen2023monogaussianavatar, xu2024gaussian, gaussianavatars, zhao2024psavatar} for avatar reconstruction. Unlike traditional mesh-based techniques, NeRF- and 3DGS-based methods can generate 3D-consistent outputs while capturing fine details that go beyond surface-level geometry.
However, these methods often require extensive frontal video data of subjects for reconstruction, and some involve optimization processes during inference that are both time-intensive and computationally demanding. To simplify input data, studies have proposed reconstructing head avatars from single-image~\cite{deng2024portrait4dv1, deng2024portrait4dv2} or multi-view image~\cite{zheng2024headgap, gaussianavatars} inputs. Nevertheless, these methods still lack robust rendering capabilities when handling unseen viewpoints.


In this paper, we present \textbf{S}-\textbf{Avatar} (\textbf{S}ingle-image-based 3DGS \textbf{Avatar}), a method for one-shot 3D head avatar reconstruction. Unlike prior neural rendering based approaches that directly optimize 3D Gaussian Splatting or NeRF representations from limited input views, which often results in degraded rendering quality under unseen viewpoints, our method follows a distinct philosophy. We first construct a canonical 3DGS head model guided by a diffusion based Gaussian splats generation module, and then control its deformation based on facial expression variations.

S-Avatar constructs an animatable 3D head avatar through three phases:  
1) initial 3D Gaussian splats generation phase, utilizing a Gaussian splats generation module, 
2) Fitting phase, where the parametric head model is aligned with initial splats to enable controllability,  
3) Binding phase, where deformation and scaling of the 3DGS are driven by variations in the FLAME model. 
We employ a diffusion-based 3DGS generation model as a Gaussian splats generation module to produce a high-resolution static 3DGS from a single image.
In the fitting phase, we fit the FLAME model to the initial splats by optimizing the FLAME parameters and transformations.
In the binding phase, we calculate a binding template, which is designed to deform Gaussian splats in response to variations in FLAME vertices.
Finally, for rendering, the splats can represent novel facial expressions and viewpoints in a photorealistic style. 
S-Avatar not only reconstructs a 3D-consistent head avatar from a single image but also achieves superior rendering quality for novel facial expressions. In addition, it enables realistic representations under extreme novel views such as side or back, which were not achievable with previous methods. These capabilities demonstrate the clear advantages of our method in the field of head avatar reconstruction.

To summarize, the contributions of our proposed method are as follows:

\begin{itemize}
    \item We introduce \textbf{S-Avatar}, a novel framework for one-shot 3D head avatar reconstruction that generates animatable avatars from a single image while maintaining 3D consistency.
    \item We propose a three-phase pipeline that incorporates initial 3D Gaussian splats generation, parametric model fitting, and a binding mechanism to achieve controllability and expressiveness.  
    \item Our method surpasses prior approaches in rendering quality and 3D consistency, producing high-fidelity avatars that generalize robustly to novel expressions and viewpoints. We further validate its applicability on in-the-wild inputs and example applications.
\end{itemize}

\section{Related Work}
Our target domain is the reconstruction of high-fidelity, controllable 3D head avatars from a single-view image. In this section, we review recent studies on 3D head avatar generation and one-shot 3D generation methods that are relevant to our proposed method, S-Avatar.

\subsection{Parametric Model Based Avatar Generation}

Reconstructing photorealistic avatars from images or video inputs has been a significant research topic in the fields of computer vision and VR/AR. Blanz et al. \cite{3dmm} proposed 3D Morphable Models (3DMMs) that represent face models by placing facial appearance and expression in a linear subspace through principal component analysis (PCA). Being able to represent facial models with low-dimensional vectors has made 3DMMs the base model for various studies in facial reconstruction. This concept was extended to full head models with the design of parametric models like the FLAME model \cite{flame}, which represents 3D heads using shape, facial expression, and pose parameters. Similarly, research into parametric body models aimed at body animation has expanded to include studies like SMPL \cite{loper2015smpl}. This work has been further integrated with the parametric hand model MANO \cite{mano} and the  parametric head model FLAME \cite{flame} to develop a holistic full-body avatar named SMPL-X \cite{SMPL-X}, which has set a standard in avatar reconstruction research.
These parametric models have been utilized in numerous studies to create realistic three-dimensional heads \cite{yu2023nofa, feng2021deca, zielonka2022mica, danvevcek2022emoca, deng2019accurate, sanyal2019learning, garrido2016reconstructionvideoface, tewari2019fmlvideoface, khakhulin2022realisticrome} and full-body avatars \cite{alldieck2018video, SMPL-X, saito2021scanimate, wang2021metaavatar, rcsmpl} based on video or image data.

\subsection{Neural Rendering Based Avatar Generation}
With neural rendering techniques such as NeRF~\cite{mildenhall2021nerf}, neural point clouds~\cite{neuralpointcloud}, and 3D Gaussian splatting (3DGS)~\cite{3dgaussiansplatting}, significant progress has been made in generating animatable 3D head avatars from monocular or multi-view videos and images, effectively capturing a diverse range of expressions and poses. Studies such as~\cite{athar2023flame, gafni2021dynamic, imavatar} have focused on creating high-fidelity 3D head avatars by leveraging the detailed expressiveness of NeRF. However, a major limitation of these approaches was the difficulty of achieving real-time rendering, which is crucial for avatars serving as a medium for user interactions in VR/AR environments. To address this, 3D Gaussian Splatting has been widely adopted in numerous studies~\cite{xiang2024flashavatar, chen2023monogaussianavatar, zhao2024psavatar, giebenhain2024npga, xu2024gaussian, gaussianavatars, song2024toward, shao2024splattingavatar} for its ability to represent objects and scenes as a set of 3D Gaussians and enable real-time rendering.

However, these methods required long monocular or multi-view video sequences to capture a diverse range of expressions and viewpoints in order to train the radiance field or the Gaussian splat set effectively. Consequently, many studies~\cite{zheng2024headgap, buhler2023preface, yang2024vrmm, li2024uravatar, yu2024one2avatar, chen2024morphable, cao2022authentic, yang2026ofera, song2026vrgaussianavatar} have shifted their focus toward reducing the number of input data required for reconstruction. Few-shot input methods have shown promising results in simplifying the input process for personalized avatar creation. However, these methods still require multi-view images of a single subject, which remains a barrier to effortless avatar generation. Additionally, some of these studies exhibit limited 3D consistency when rendering unseen viewpoints.
To overcome these limitations, we propose a 3DGS-based head avatar generation and control method that enables robust rendering from a single image, including under unseen or extreme viewpoints such as side and back views.


\begin{figure*} [ht]
  \includegraphics[width=\textwidth]{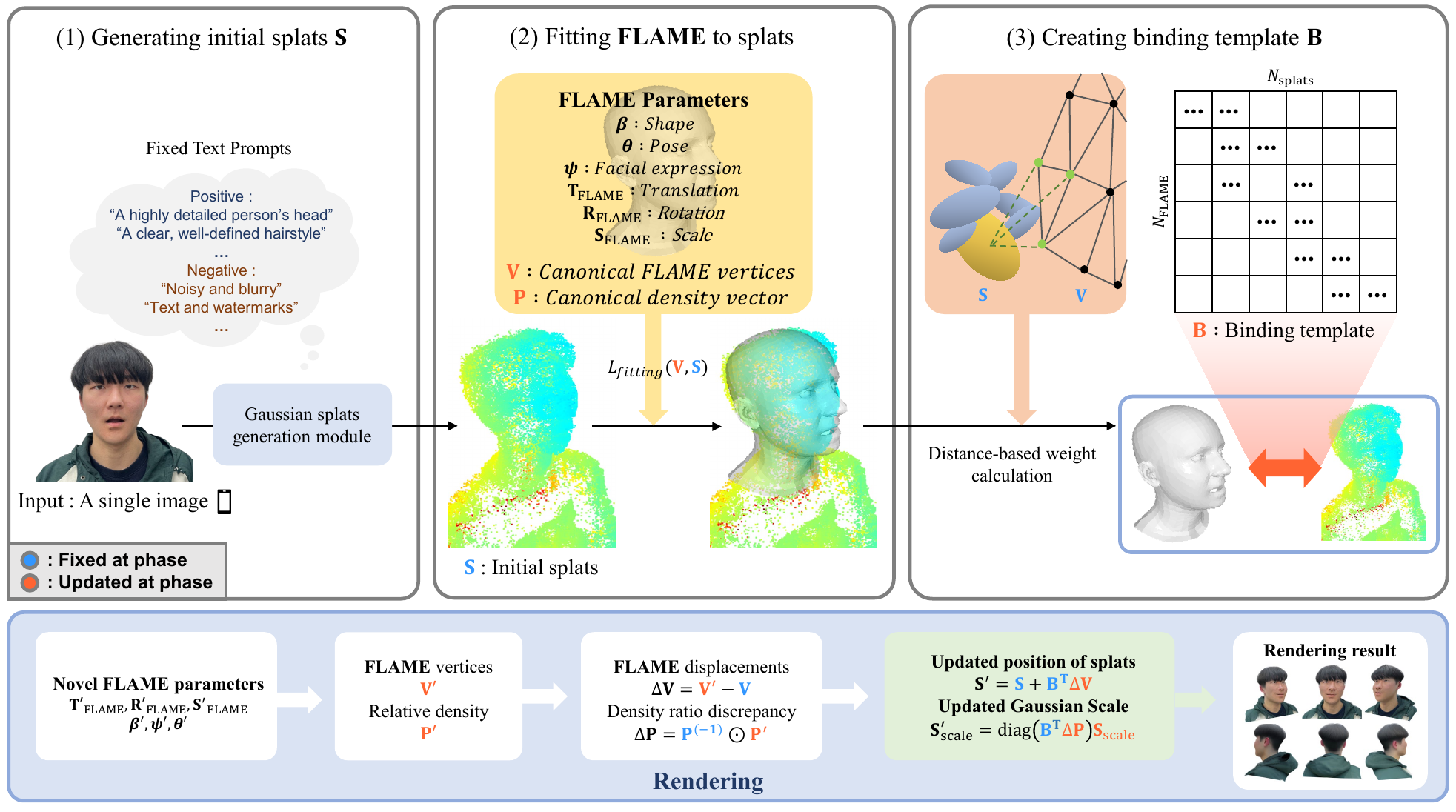}
  \caption{
    System diagram of the proposed method, \textbf{S-Avatar}. Our method reconstructs a photorealistic head avatar from a single image input through a three-step modeling process: (1) Generating initial splats using Gaussian splats generation module, (2) Fitting the FLAME model to the generated Gaussian splats, and (3) Creating a binding template. During the rendering phase, the generated avatar can display various facial expressions and be rendered from novel viewpoints.
  }
  \label{fig:overview}
\end{figure*}

\subsection{One-shot 3D Model Generation for Head Avatars}

While traditional head avatar generation methods rely on multi-view data or long video sequences, recent studies~\cite{zheng2024headgap, deng2024portrait4dv1, deng2024portrait4dv2, tran2024voodoo3d, khakhulin2022realisticrome, chu2024gpavatar, li2023onehide, li2023generalizablegoha, he2025lam, moon2025geoavatar, guo2025sega, zhang2025hravatar, zhou2025zero} have increasingly focused on simplifying the input to a single image.
One-shot head avatar generation has gained attention due to its potential for accessible and user-friendly avatar personalization.
Despite this progress, many existing approaches struggle to represent novel expressions and to maintain high-quality reconstruction of unobserved regions, such as the backside of the head.

Given that multi-view consistency is also crucial in image-to-3D tasks, recent studies have begun addressing this challenge as well. In particular, various approaches have integrated image-to-3D models with 3DGS techniques to meet the demands of geometric consistency and rendering quality.
First, optimization-based approaches~\cite{tang2023dreamgaussian, yi2024gaussiandreamer, chen2024text} utilize Score Distillation Sampling to iteratively refine 3D models. While these methods yield high-fidelity results, their computational process is significantly slow. 3D-native diffusion methods~\cite{mu2024gsd, he2024gvgen, zhang2024gaussiancube} train diffusion models directly on 3D representations, thereby enhancing multi-view consistency. However, these methods encounter challenges in scaling training datasets to achieve broader generalization~\cite{chen2024survey}. In contrast, reconstruction-based approaches~\cite{melas20243d, lee2024vividdream}, including Large multi-view Gaussian Model (LGM)~\cite{tang2024lgm}, employ pretrained multi-view diffusion models to reframe the task as multi-view reconstruction. This strategy allows for rapid image generation and easier dataset scalability, facilitating improved generalization. Furthermore, LGM specifically utilizes an asymmetric U-Net structure to preserve multi-view consistency effectively.

Building on these observations, we introduce a novel framework composed of a fitting module and a binding module, leveraging the FLAME model in conjunction with initial Gaussian splats generated by a diffusion-based module. Our approach achieves one-shot, high fidelity 3D head avatar reconstruction with enhanced realism and controllability.

\section{Method}
\autoref{fig:overview} presents the overall pipeline of S-Avatar, which takes a single image as input and processes it through the following stages:
\begin{itemize}
    \item[(1)] Generate the initial splats from a single RGB image using a diffusion-based 3DGS generation module.
    \item[(2)] Fit the FLAME head model to the initial splats to establish geometric alignment and controllability.
    \item[(3)] Compute a binding template that drives Gaussian splat deformations according to FLAME variations.
\end{itemize}
By generating the 3D Gaussian splats from a single-view image using Gaussian splats generation module and binding it to a controllable parametric head model, our method achieves photo-realistic rendering results with fast reconstruction and supports animatable avatars. Compared to conventional methods that rely on dense multi-view inputs or lengthy optimization procedures, S-Avatar simplifies the data acquisition process while maintaining high-quality avatar reconstruction.

\subsection{Preliminary}
~\\
\textbf{3D Gaussian Splatting.   }
3D Gaussian Splatting (3DGS) \cite{3dgaussiansplatting} is a method to reconstruct static scenes using 3D Gaussian splats as a neural primitive based on images and camera parameters. Each scene component is represented as a set of Gaussian splats on 3D dimension, defined by the following equation:
\begin{equation}
G(x) = e^{-\frac{1}{2}(x-\mu)^T\Sigma^{-1}(x-\mu)}
\label{eq:gaussian}
\end{equation}
\begin{equation}
\Sigma = RS S^T R^T
\label{eq:covariance}
\end{equation}
In \autoref{eq:gaussian}, $G(x)$ describes a Gaussian splat, where $(x-\mu)$ measures the displacement from the center point $\mu$, and $\Sigma^{-1}$ is the inverse of the covariance matrix $\Sigma$. This matrix dictates the spread and orientation of the splat in 3D space.
To guarantee stability in gradient descent processes, Kerbl et al.~\cite{3dgaussiansplatting} introduced a parametric ellipse model that replaces direct optimization of the covariance matrix, as demonstrated in \autoref{eq:covariance}. The ellipse is defined by a scaling vector $s \in \mathbb{R}^3$, position vector $\mu \in \mathbb{R}^3$ and a quaternion $q \in \mathbb{R}^4$.
The position vector $\mu$, explicitly defined in the Gaussian splat as a neural primitive, facilitates intuitive representation of dynamic facial expressions by enabling straightforward adjustments to the splat locations.

During the rendering process, the color of each splat is determined by its spherical harmonics and the view direction. The rendering result for each pixel $C$ is computed by accumulating contributions from overlapping Gaussian splats.

\begin{equation}
C = \sum_{i=1}^{N} c_i \alpha_i' \prod_{j=1}^{i-1} (1 - \alpha_j')
\label{eq:color_blending}
\end{equation}
In \autoref{eq:color_blending}, $c_i$ represents the color value of each splat, and $\alpha'$ is a blending weight derived by evaluating the projection of the Gaussian splats, which is calculated using a per-splat opacity $\alpha$. To maintain the correct visual order, Gaussian splats are sorted by depth before blending.
The adoption of 3DGS in our method facilitates rapid avatar reconstruction and rendering. Moreover, we adjust splats' positions and scales in response to variations of the parametric head model, specifically the FLAME model to represent various facial expressions.

~\\
\textbf{FLAME head model.   }
The core structure of the FLAME model is shown in \autoref{eq:FLAME}.

\begin{equation}
    M(\beta, \theta, \psi) = LBS(T_P(\beta, \theta, \psi), J(\beta, \theta), \beta, W)
    \label{eq:FLAME}
\end{equation}
\begin{equation}
    T_P(\beta, \theta, \psi) = \overline{T} + B_S(\beta; S) + B_P(\theta; P) + B_E(\psi; E)
    \label{eq:templateFLAME}
\end{equation}
\autoref{eq:FLAME} presents the function $M$ that computes the mesh output, utilizing the FLAME parameters to transform the model. Here, \(\beta\), \(\theta\), and \(\psi\) represent the shape, pose, and expression parameters, respectively. The function \(LBS(\cdot)\) denotes the Linear Blend Skinning, which applies deformations to the template mesh based on these parameters. \(J(\cdot)\) calculates the transformations for each joint in the hierarchy, and \(W\) indicates the per-vertex skinning weights for smooth transitions. 
\autoref{eq:templateFLAME} shows that the customized neutral template mesh \(T_{P}\) is adjusted by blendshapes for shape \(B_S(\cdot)\), pose \(B_P(\cdot)\), and expressions \(B_E(\cdot)\). These blendshapes, derived from the bases \(S\) (Shape), \(P\) (Pose), and \(E\) (Expression), adapt dynamically to the parameters \(\beta\), \(\theta\), and \(\psi\), facilitating detailed and realistic facial animations. During rendering phase, the system takes novel expression and pose parameters to compute FLAME vertices with personalized shape parameters.

\subsection{Initial 3D Gaussian Splats Generation}
\label{sec:lgm_generation}

To generate a 3D head representation from a single image, our method utilizes a dedicated Gaussian splats generation module. This feed-forward network predicts a set of 3D Gaussian splats through a multi-view image fusion process, starting from a single input image.

At inference time, the module first leverages a pre-trained multi-view diffusion model, ImageDream~\cite{wang2023imagedream}, to synthesize four view-consistent images, each guided by a distinct orthogonal azimuth angle at a fixed elevation. These synthesized views, together with their corresponding camera parameters, are subsequently fed into an asymmetric U-Net backbone to predict 3D Gaussian representations. Each input image is paired with its camera pose and embedded into a feature map that encodes both visual and geometric information. The asymmetric U-Net with cross-view self-attention processes these multi-view feature maps and generates four sets of 14-channel maps, where each pixel represents the attributes of a 3D Gaussian splat. These outputs are then fused into a unified set, which we define as the initial splats $\mathbf{S}$.
This generation module is trained under a multi-view reconstruction setting, supervised by differentiable rendering using synthetic images generated from 3D datasets. This design enables the reconstruction of dense and photorealistic 3D Gaussian representations without requiring iterative optimization. We adopt the specific architecture for this module from the Large Multi-view Gaussian Model (LGM)~\cite{tang2024lgm}.

The resulting initial splats $\mathbf{S}$ provide a high-quality and efficiently rendered representation of the subject, serving as the foundation for downstream fitting and animation in the S-Avatar pipeline. To generate robust 3DGS results, we carefully designed both positive and negative text prompts. These prompts serve as additional inputs to guide the generation of initial splats. Detailed descriptions of the text prompts are provided in the supplemental material.

\begin{figure}[t]
\centering
 \includegraphics[width=0.9\columnwidth]{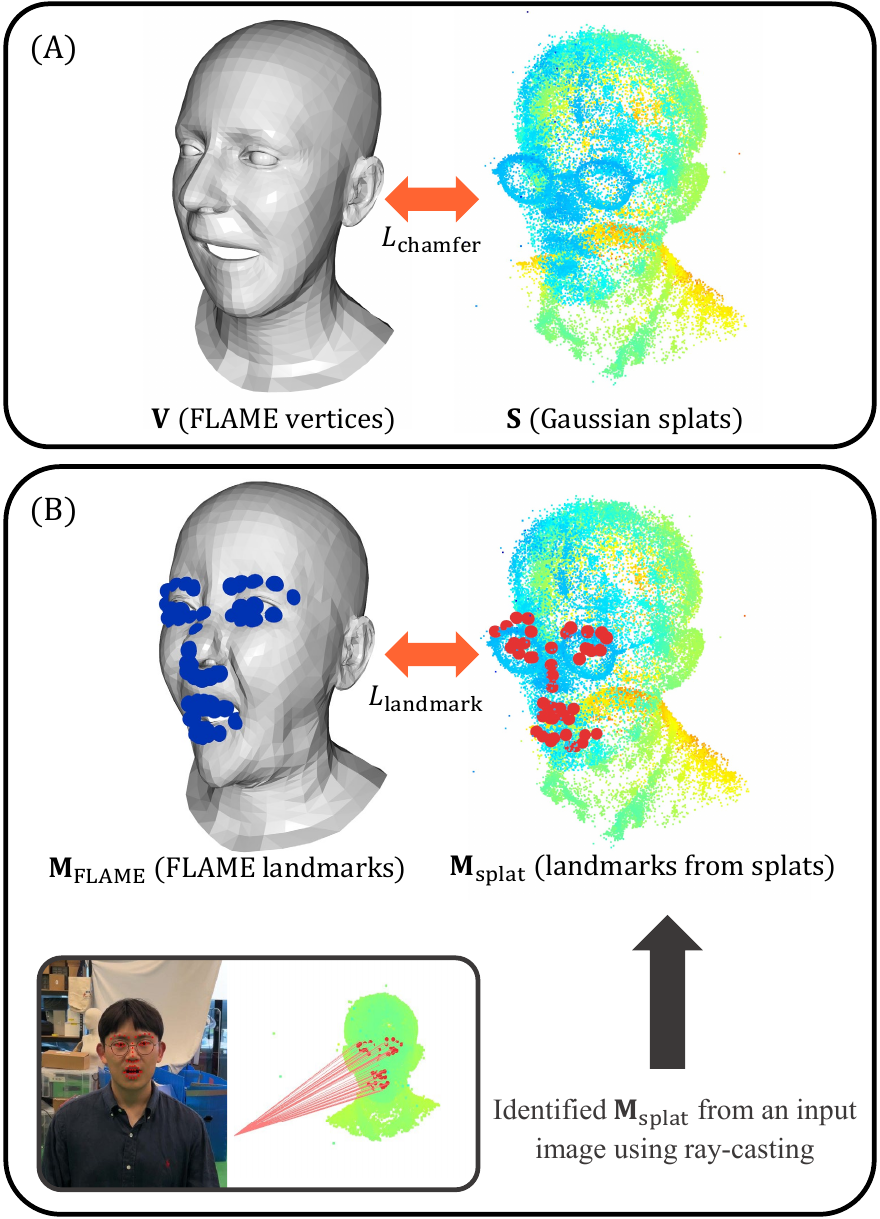}
 \caption{
  The fitting losses of S-Avatar. (A) shows the chamfer distance loss $L_{\text{chamfer}}$ between FLAME vertices $\mathbf{V}$ and Gaussian splats $\mathbf{S}$, and (B) displays the landmark loss $L_{\text{landmark}}$, which is the Euclidean distance between FLAME landmarks $\mathbf{M}_{\text{FLAME}}$ and landmarks derived from splats, $\mathbf{M}_{\text{splat}}$. The landmark loss $L_{\text{landmark}}$ is obtained through ray-casting onto the Gaussian splat set $\mathbf{S}$ using 2D landmarks from the input image.
 }
 \label{fig:fitting}
\end{figure}

\vspace{0.4\baselineskip}

\subsection{Fitting FLAME Model to 3D Gaussian Splats}
\label{subsec:fitting}

To align the FLAME head model with the initial splats $\mathbf{S}$, we optimize FLAME parameters consisting of translation $\mathbf{T}_{\text{FLAME}} \in \mathbb{R}^3$, rotation $\mathbf{R}_{\text{FLAME}} \in \mathbb{R}^3$, scale $\mathbf{S}_{\text{FLAME}} \in \mathbb{R}$, shape $\boldsymbol{\beta} \in \mathbb{R}^{300}$, expression $\boldsymbol{\psi} \in \mathbb{R}^{100}$, and pose $\boldsymbol{\theta} \in \mathbb{R}^{6}$ parameters through the gradient-based optimization method Adam \cite{adam}.
The FLAME head model is deformed based on the canonical template mesh \(\bar{T}\). Therefore, if the vertices corresponding to the key facial landmarks in the template mesh are known, the landmark information can be treated as a function that outputs not only the vertices but also the 3D landmarks. We defined the function \(F\), which takes the FLAME parameters along with the transformation and scale information as input, to output the transformed FLAME model's vertices $\mathbf{V}$ and the 3D landmark \(\mathbf{M}_{\text{FLAME}}\) as follows.
\begin{equation}
    \{ \mathbf{V}, \mathbf{M}_{\text{FLAME}} \} = F(M(\beta,\, \theta,\, \psi),\, \mathbf{T}_{\text{FLAME}},\, \mathbf{R}_{\text{FLAME}},\, \mathbf{S}_{\text{FLAME}} )
\label{eq:FLAME_new}
\end{equation}
The FLAME vertices $\mathbf{V}$ and FLAME landmarks $\mathbf{M}_{\text{FLAME}}$ are used to compute the discrepancy between the FLAME model and the initial splats during the optimization of the FLAME parameters.

The objective function, denoted as $L_{fitting}(\mathbf{V}, \mathbf{S})$, measures the discrepancy between the current state of the FLAME vertices $\mathbf{V}$ and the target initial splats set $\mathbf{S}$ through a Chamfer distance loss \cite{chamfer_distance} and landmark alignment loss $L_{\text{landmark}}$. 
First, the Chamfer Distance loss $L_{\text{chamfer}}$ is computed between the vertices $\mathbf{V}$ of the FLAME model and the splats in $\mathbf{S}$ as shown in \autoref{fig:fitting} (A), expressed as:
\begin{equation}
L_{\text{chamfer}} = \frac{1}{N_{\text{FLAME}}} \sum_{v \in \mathbf{V}} \min_{s \in \mathbf{S}} | v - s |^2 + \frac{1}{N_{\text{splat}}} \sum_{s \in \mathbf{S}} \min_{v \in \mathbf{V}} | s - v |^2
\label{eq:chamfer}
\end{equation}
$N_{\text{FLAME}}$ represents the number of vertices of the FLAME model $\mathbf{V}$, and $N_{\text{splat}}$ denotes the number of initial splats $\mathbf{S}$.

\noindent Second, the landmark alignment loss $L_{\text{landmark}}$ is computed as:
\begin{equation}
L_{\text{landmark}} = \sum_{i=1}^{N_{\text{landmark}}} | \mathbf{M}_{\text{FLAME}, i} - \mathbf{M}_{\text{splat}, i} |^2
\label{eq:landmarkloss}
\end{equation}
where $\mathbf{M}_{\text{FLAME}, i}$ represents the $i$-th element of FLAME landmark $\mathbf{M}_{\text{FLAME}}$, $\mathbf{M}_{\text{splat}, i}$ corresponds to the $i$-th facial landmark predicted from the Gaussian splat set $\mathbf{S}$, and $N_{\text{landmark}}$ denotes the total number of landmarks considered in the alignment process.
We obtain the 3D facial landmarks $\mathbf{M}_{\text{FLAME}, i}$ from the FLAME model parameterized accordingly.
$\mathbf{M}_{\text{splat}, i}$ denotes the 3D Gaussian splat corresponding to the $i$-th 2D facial landmark. It is determined by performing ray-casting from the 2D landmark in the input image to locate its corresponding splat in the 3DGS, as illustrated in \autoref{fig:fitting} (B).
Our loss function measures the squared Euclidean distance between each pair of corresponding landmarks, effectively quantifying the discrepancy in their positions. The metric is essential for ensuring that the FLAME model accurately captures the geometric features of the 3D data, facilitating a precise alignment between the model and the observed 3D landmarks.

The full objective loss $L_{fitting}$ is computed with a weighted combination of losses. 
\begin{equation}
    L_{fitting}(\mathbf{V}, \mathbf{S}) = \lambda_{c}L_{\text{chamfer}} + \lambda_{l}L_{\text{landmark}}
    \label{eq:fitting_loss}
\end{equation}
$\lambda_c$ and $\lambda_l$ are weighting coefficients that balance the influence of losses, which are set as $\lambda_c = 0.01$ and $\lambda_l = 0.99$, respectively.

\subsection{Binding Gaussian Splats to FLAME}

\begin{figure}[t]
 \centering
 \includegraphics[width=0.9\columnwidth]{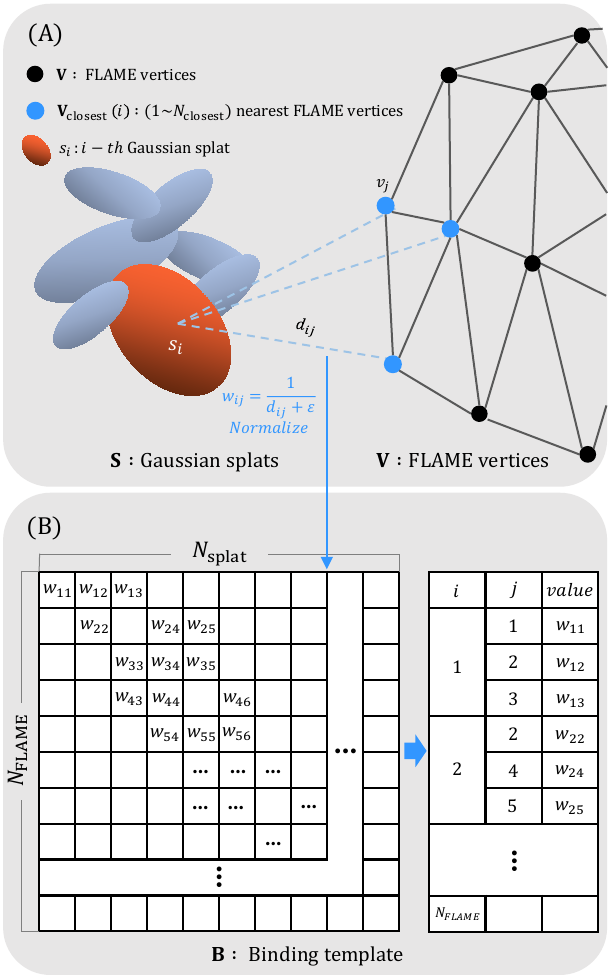}
 \caption{
 The binding process of S-Avatar.
 (A) shows the process of finding the $(1 \ldots N_{\text{closest}})$ closest FLAME vertices to the $i$-th Gaussian splat $s_i$, calculating the distance $d_{ij}$ and the weight $w_{ij}$. (B) demonstrates how these weights are stored with COO format matrix.
 }
 \label{fig:binding}
\end{figure}

The final step for animating initial splats $\mathbf{S}$ for novel facial expressions is binding them to the FLAME model's vertices $\mathbf{V}$. In our method, each Gaussian splat, the primitive representing the head, is designed to adapt and move in response to the movements of the nearby FLAME vertices through the proposed binding template. As these two sets are affixed with each corresponding element after the generating process, we can transform and scale the splats based on changes in FLAME vertices using inverse distance weighting which is defined as the following equations.

First, we calculate the Euclidean distance between the $i$-th splat $s_i$ and the $j$-th FLAME vertex $v_j$.
\begin{equation}
d_{ij} = \| s_i - v_j \|_2
\label{eq:euclidean_distance}
\end{equation}
Next, we compute weights $w_{ij}$ as the inverse of the distance between splat and FLAME vertex, adding a small constant $\epsilon$ to avoid division by zero.
\begin{equation}
w_{ij} = \frac{1}{d_{ij} + \epsilon}
\label{eq:inverse_distance_weight}
\end{equation}
The weights are normalized for each splat to ensure the sum of weights across its \( N_{\text{closest}} \) number of nearest vertices, which are collectively referred to as \( \mathbf{V}_{\text{closest}} \). 
To select the $N_{\text{closest}}$ closest points, we make use of the quick sort algorithm \cite{quicksort}.
The matrix \( \mathbf{V}_{\text{closest}}(i) \) has the position information of the nearest FLAME vertices to $i$-th Gaussian splat.
\begin{equation}
w_{ij}^\prime = \frac{w_{ij}}{\sum_{k \in \mathbf{V}_{\text{closest}}(i)} w_{ik}}
\label{eq:weight_normalization}
\end{equation}
Then the binding template $\mathbf{B} \in \mathbb{R}^{N_{\text{FLAME}} \times N_{\text{splat}}}$ is constructed with a set of $w'_{ij}$ for $i = 1, \ldots, N_{\text{FLAME}}$ and $j = 1, \ldots, N_{\text{splat}}$, where each $w'_{ij}$ represents the normalized weight between the $i$-th FLAME vertex and the $j$-th splat. It is used to update the position of each splat by calculating the weighted average of the displacements of its vertices in \( \mathbf{V}_{\text{closest}} \).
 

At the rendering phase, we compute the displacement of the splats, $\Delta\mathbf{S}$, by applying the transposed binding matrix $\mathbf{B}^T$ to the vertex displacements.
\begin{equation}
    \Delta\mathbf{S} = \mathbf{B}^T (\mathbf{V'} - \mathbf{V})
    \label{eq:splat_displacement}
\end{equation}
Here, each element of $\Delta\mathbf{S} \in \mathbb{R}^{N_{\text{splat}} \times 3}$ corresponds to the displacement of a splat, computed as the weighted average of displacements of its closest FLAME vertices. The term $(\mathbf{V'} - \mathbf{V})$ represents the matrix of vertex displacements between $\mathbf{V}$ from \autoref{subsec:fitting} and $\mathbf{V'}$ from novel FLAME parameters.
Finally, the updated positions $\mathbf{S'}$ are obtained by adding this displacement to the original positions of the initial splats $\mathbf{S}$.
\begin{equation}
\mathbf{S'} = \mathbf{S} + \Delta\mathbf{S}
\label{eq:update_positions}
\end{equation}

Furthermore, to account for the stretching and compressing of the facial surface during expression changes, we introduce a Gaussian scale adaptation method. This technique adjusts the scale of each splat based on the change in the relative density of the FLAME mesh vertices it is bound to.
Specifically, we define the relative density $P_j$ for each vertex $v_j$ in the canonical FLAME $\mathbf{V}$. This value is the inverse of the average length of its connected edges, calculated using $N(j)$, the set of neighboring vertex indices.
\begin{equation}
    P_j = \left( \frac{1}{|N(j)|} \sum_{k \in N(j)} ||v_k - v_j||_2 \right)^{-1}
\end{equation}
The canonical density vector $\mathbf{P} \in \mathbb{R}^{N_{\text{FLAME}}}$ is then constructed by collecting the relative density $P_j$ for all vertices.

Using the binding template $\mathbf{B} \in \mathbb{R}^{N_{\text{FLAME}} \times N_{\text{splat}}}$, we compute the scale factor vector $\mathbf{\sigma} \in \mathbb{R}^{N_{\text{splat}}}$ for all splats in a single operation. This is achieved by multiplying the transpose of the binding matrix with the element-wise product of the element-wise reciprocal of the canonical density vector $\mathbf{P}$ and the updated density vector $\mathbf{P'}$ calculated from novel FLAME parameters.
\begin{equation}
    \mathbf{\sigma} = \mathbf{B}^T (\mathbf{P}^{(-1)} \odot \mathbf{P'})
\end{equation}
Here, $\mathbf{P}^{(-1)}$ denotes the element-wise reciprocal of the vector $\mathbf{P}$, and $\odot$ represents element-wise multiplication (the Hadamard product).
Finally, the updated scale matrix $\mathbf{S'}_{\text{scale}} \in \mathbb{R}^{N_{\text{splat}} \times 3}$ is obtained by modulating the initial scales $\mathbf{S}_{\text{scale}}$ with a diagonal matrix formed from $\mathbf{\sigma}$.
\begin{equation}
    \mathbf{S'}_{\text{scale}} = \text{diag}(\mathbf{\sigma}) \mathbf{S}_{\text{scale}}
\end{equation}
This ensures that splats in regions of the mesh that stretch become larger, and splats in regions that compress become smaller, preserving the apparent surface density.

Considering that each row of the binding template contains only \( N_{\text{closest}} \) non-zero elements, we always compute the sparse matrix with large dimensions. This structure allows us to store and process efficiently \( \mathbf{B} \) using the COO (Coordinate List) format \cite{COO}. By adopting the COO format, we can perform operations on \( \mathbf{B} \) more effectively \cite{COOevaluate}, minimizing memory usage and optimizing computational efficiency.

\begin{figure*}[!ht]
\centering
\includegraphics[width=0.980\textwidth,keepaspectratio]{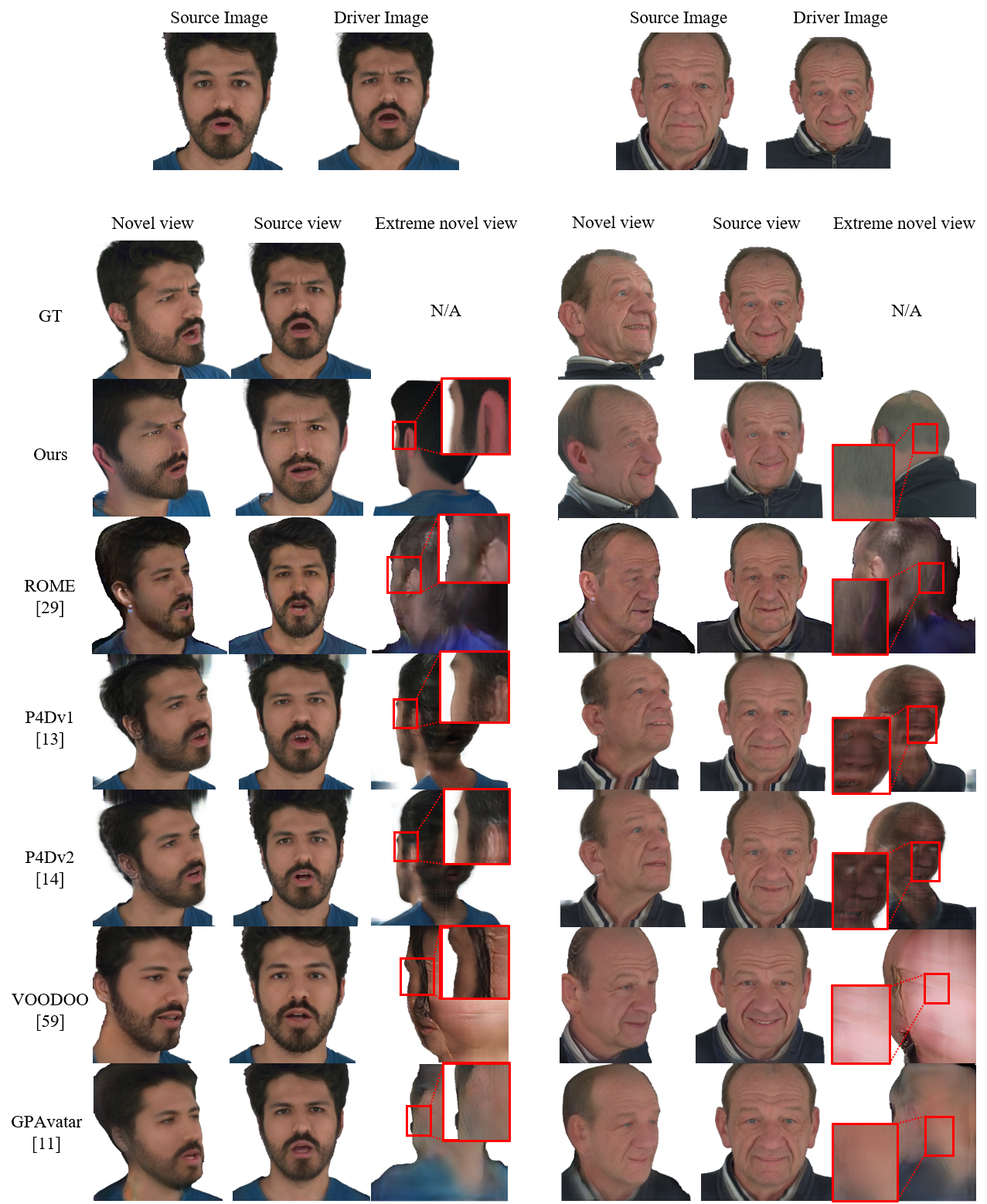}
\caption{
    Multi-view self-reenactment rendering results for S-Avatar and baseline methods. Both our method and the baseline generate avatars using a source image, and they utilize expressions from a driver image to depict new facial expressions. Our method demonstrates more realistic facial expression rendering and a closer resemblance to the person in the source image compared to the baseline. Additionally, it maintains realistic expression rendering under extreme novel views, outperforming other methods. The IDs correspond to the NeRSemble dataset~\cite{kirschstein2023nersemble}.}
\label{fig:qualitative2}
\end{figure*}

\begin{figure*}[!ht]
\centering
\includegraphics[width=\textwidth,keepaspectratio]{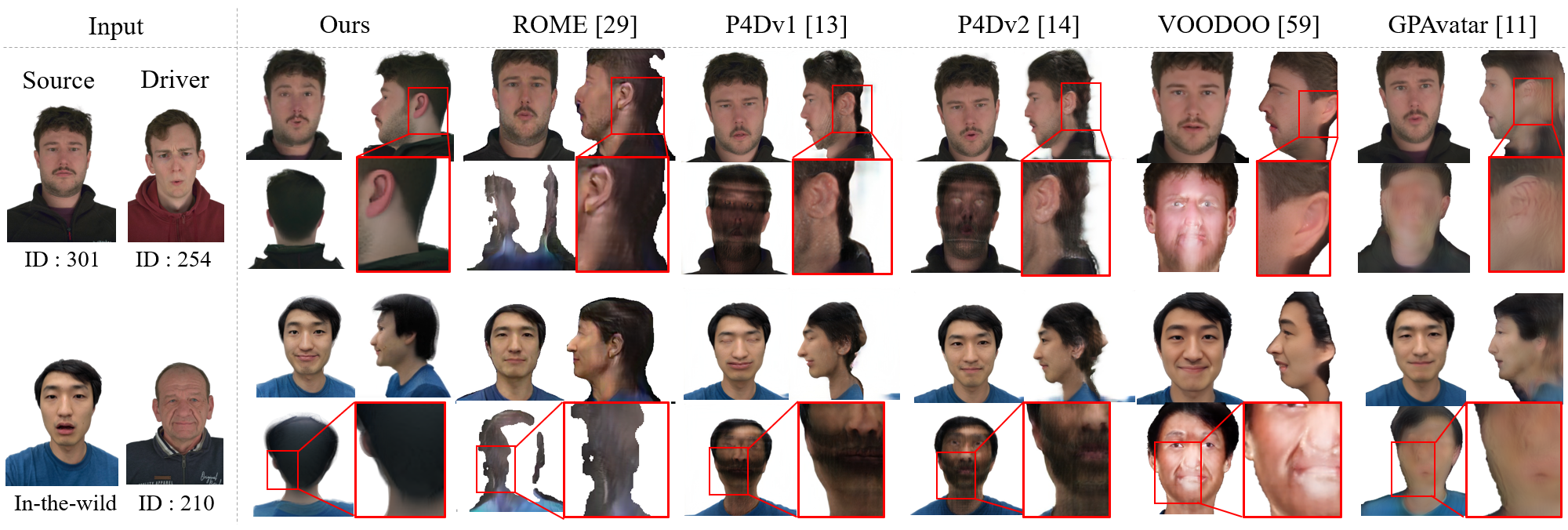}
\caption{
    Qualitative comparison of cross-reenactment results. Each avatar is reconstructed from a source image and driven by the expression of a driver image. Our method better preserves identity and more accurately transfers novel expressions, even under challenging views.
}
\label{fig:qualitative1}
\end{figure*}

\section{Experiments and Results}

\subsection{Experimental Setup}
~\\
\textbf{Baselines.   }
We compared our approach to several baselines for generating animatable 3D head avatars from a single image, through quantitative and qualitative evaluation. Each baseline is a state-of-the-art method employing different techniques for one-shot head avatar generation: mesh-based modeling~\cite{khakhulin2022realisticrome}, neural self-supervised disentanglement~\cite{tran2024voodoo3d}, GAN-based approaches~\cite{deng2024portrait4dv1}, and pseudo multi-view data methods~\cite{deng2024portrait4dv2}.

~\\
\textbf{Datasets.   }
To evaluate the expressive and view-consistent capabilities of head avatars, we utilize the NeRSemble dataset~\cite{kirschstein2023nersemble} for both quantitative and qualitative comparison. NeRSemble consists of 164 subjects captured from 16 camera viewpoints with diverse facial expressions. Among these, we selected 45 subjects as the test set to ensure fair quantitative evaluation across methods. Since each baseline method has different avatar reconstruction scopes and employs different facial tracking strategies, the resulting face crop ranges vary. Therefore, to ensure fairness, we evaluate each method by cropping both the ground-truth image and the rendered result according to the method's own face cropping strategy, so that the comparison is performed within the same facial region. This avoids penalizing methods due to mismatched crop boundaries.
For the quantitative evaluation, we used all 16 camera viewpoints provided by NeRSemble. For the qualitative comparison, we additionally present results from side and back views, which are significantly different from the input image's viewpoint.

~\\
\textbf{Metrics.   }
To quantitatively compare our method with the baselines, we employed standard image similarity metrics: LPIPS~\cite{lpips}, PSNR~\cite{psnr}, and SSIM~\cite{ssim}.
LPIPS (Learned Perceptual Image Patch Similarity) is a learned metric that quantifies image differences based on deep features extracted from neural networks, aligning more closely with human visual perception.
PSNR (Peak Signal-to-Noise Ratio) measures the pixel-wise fidelity between the rendered image and the ground truth, with higher PSNR values indicating better image quality.
SSIM (Structural Similarity Index) evaluates perceptual similarity between two images by comparing their structural information, such as luminance and contrast.

~\\
\textbf{Implementation details.   }
All processing for both our method and the baselines was conducted on an RTX 3090 GPU with 24GB of VRAM.  
During the fitting process of the FLAME model to the initial splats, we optimized the parameters over 60 epochs. 
We set the learning rates as follows: 0.005 for the shape parameters, 0.0005 for the expression, pose, neck, and jaw parameters, and 0.01 for the global transformation (translation, rotation, and scaling).  
For the Adam optimizer, we used $\beta_1 = 0.9$, $\beta_2 = 0.999$, and $\epsilon = 1\mathrm{e}{-8}$.

\subsection{Quantitative Results}

\newcommand{\mybox}[2]{%
  {\setlength{\fboxsep}{2pt}\colorbox[HTML]{#1}{#2}}%
}

\begin{table}[]
\centering
\renewcommand{\arraystretch}{1.3} 
\begin{tabular}{llccc}
\hline
\textbf{Method} & \multicolumn{1}{c}{\textbf{LPIPS $\downarrow$}} & \multicolumn{1}{c}{\textbf{PSNR $\uparrow$}} & \multicolumn{1}{c}{\textbf{SSIM $\uparrow$}} \\ \hline
ROME\cite{khakhulin2022realisticrome} & 0.286 & \mybox{C0C0C0}{15.45} & \mybox{C0C0C0}{0.796} \\
P4Dv1\cite{deng2024portrait4dv1} & \mybox{ED9B5D}{0.273} & 14.52 & 0.752 \\
P4Dv2\cite{deng2024portrait4dv2} & \mybox{C0C0C0}{0.269} & 14.53 & 0.757 \\
VOODOO\cite{tran2024voodoo3d} & 0.310 & \mybox{ED9B5D}{14.59} & 0.752 \\
GPAvatar\cite{chu2024gpavatar} & 0.294 & 13.83 & \mybox{ED9B5D}{0.775} \\
Ours & \mybox{FFFC9E}{0.258} & \mybox{FFFC9E}{16.10} & \mybox{FFFC9E}{0.826} \\ \hline
\end{tabular}
\caption{Quantitative evaluations with state-of-the-art methods on the NeRSemble~\cite{kirschstein2023nersemble} dataset. Color codes denote the \mybox{FFFC9E}{\strut first}, \mybox{C0C0C0}{\strut second}, and \mybox{ED9B5D}{\strut third} rankings, respectively. Our approach outperforms current methods in LPIPS~\cite{lpips} and PSNR~\cite{psnr}, while yielding comparable results in SSIM~\cite{ssim}.}
\label{quantitative}
\end{table}


To eliminate the influence of background on avatar reconstruction performance, we preprocessed the source image using MODNet~\cite{MODNet}. Each method takes the source image as input and produces rendered results under new facial expressions and novel views. To evaluate the rendering results, we extracted expression and pose data from a driver image that represents a new expression, and used this data for rendering. Unlike some baselines that require a driver image to represent the driving expression, our method can animate the avatar using only FLAME parameters and a novel camera matrix as inputs, thus requiring less information. We employed a tracker based on MICA~\cite{zielonka2022mica} to extract FLAME parameters from the test datasets. Some methods aim to reconstruct the full face, while others focus on synthesizing new poses for portrait views. Therefore, for a fair comparison, we cropped the region around the face based on the tracking results and compared it with the corresponding ground truth to obtain quantitative results. Our method and all baselines were rendered and evaluated across 16 views of the NeRSemble dataset.


As shown in \autoref{quantitative}, our method outperformed the baselines in LPIPS, PSNR, and SSIM. The superior LPIPS score indicates higher perceptual similarity, as it is based on features extracted from deep networks. Furthermore, the higher PSNR (Peak Signal-to-Noise Ratio) and SSIM (Structural Similarity Index) values demonstrate that our method experiences less quality degradation and better preserves structural details in novel expressions and views compared to the baselines.

\begin{figure}[t]
\centering
\includegraphics[width=\columnwidth,keepaspectratio]{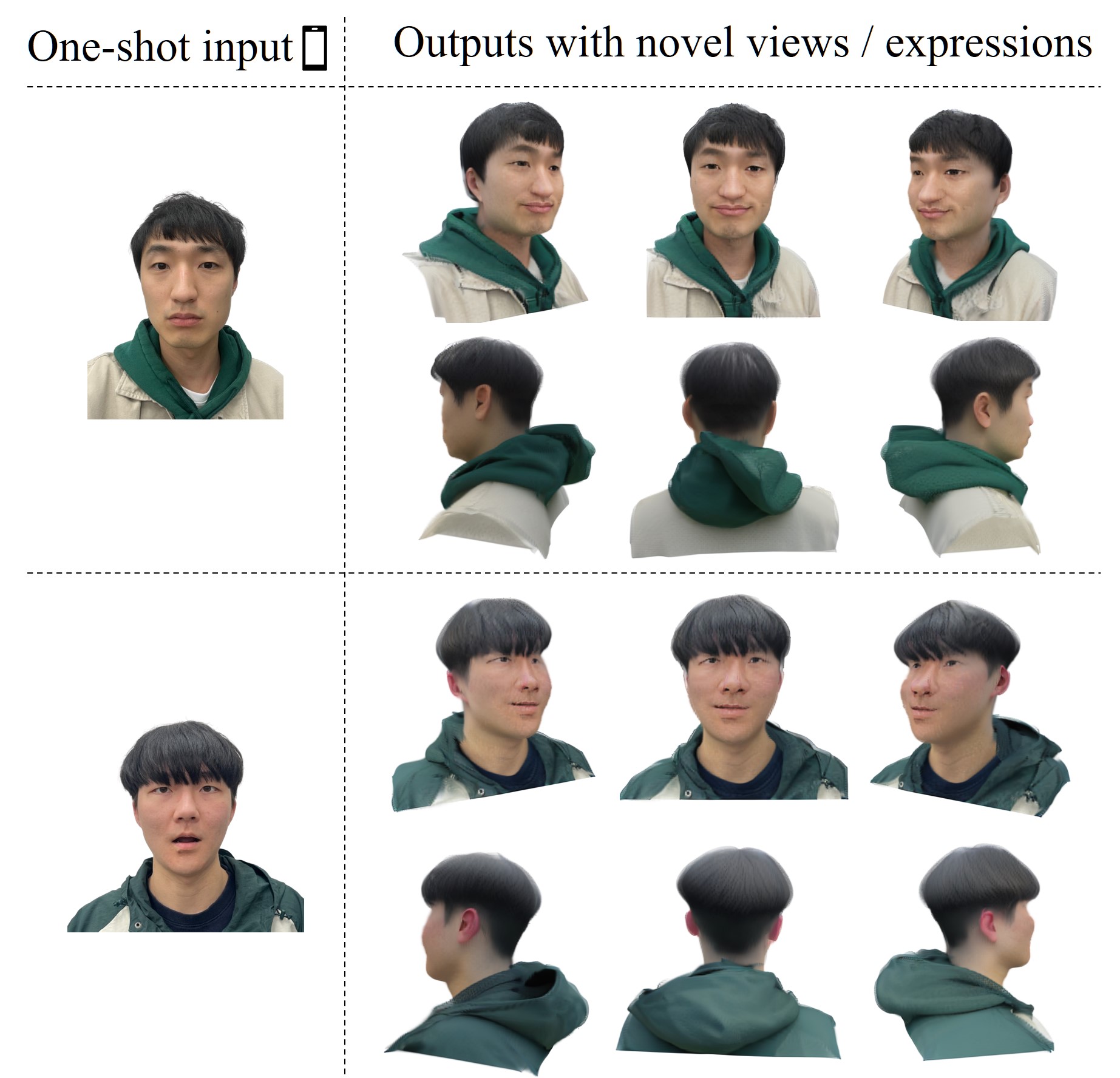}
\caption{
Head avatar generation results from a one-shot, in-the-wild image. The left shows a single RGB image captured using a commercial smartphone, while the right displays rendering results featuring novel views and facial expressions.
}
\label{fig:qualitative3}
\end{figure}

\subsection{Qualitative Results}

\autoref{fig:qualitative2} provides a qualitative comparison using rendering results from novel views not used during avatar generation. Our method not only allows rendering of novel expressions from various views but also better preserves the identity captured from the source. In particular, under extreme novel view conditions, our method consistently outperforms the baselines in rendering quality. As highlighted in the boxed regions of \autoref{fig:qualitative2}, it accurately reconstructs head details such as the ears and the back of the head, which are often distorted or missing in the baseline results. This demonstrates the robustness of our approach not only in representing novel facial expressions but also in preserving geometric consistency from challenging viewpoints.

\vspace{10pt}

\autoref{fig:qualitative1} shows a qualitative comparison of cross-reenactment results, where the source and driver images come from different identities. Each head avatar is reconstructed from the source image and animated using the facial expression extracted from the driver image. Compared to the baselines, our method more faithfully reproduces the intended expressions and maintains high rendering fidelity, especially for challenging novel expressions and views. Notably, our method successfully renders detailed geometry around the ears and the back of the head, which are often missing or corrupted in other methods.


To demonstrate the applicability of our method in real-world scenarios, we reconstruct avatars from RGB images captured by a commercial smartphone. As shown in \autoref{fig:qualitative3}, the reconstructed avatars from in-the-wild images can represent dynamic facial expressions in a photo-realistic manner.
We further demonstrate the practical utility of our method through both an interactive viewer and a real-time VR demo. The interactive viewer enables real-time control of FLAME parameters, while the VR demo translates internal sensor data from a commercial VR HMD into FLAME expression parameters using OFERA~\cite{yang2026ofera}, enabling dynamic head avatar animation in real time. In addition to faithfully reflecting facial expressions from the frontal view, our method supports robust rendering from unobserved viewpoints. Additional qualitative examples and demo results are provided in the supplemental materials.


\subsection{Ablation Studies}


\begin{table}[t]
\centering
\renewcommand{\arraystretch}{1.7}
\begin{tabular}{c|c|cccc}
\hline
\textbf{Condition} & \textbf{Quality} & \multicolumn{4}{c}{\textbf{Processing time (sec)}} \\ \hline
\multirow{2}{*}{$N_{closest}$} & \multirow{2}{*}{PSNR} & \multicolumn{3}{c}{\textbf{3DGS Generation}} & \textbf{Rendering} \\ \cline{3-6} 
 &  & LGM & fit & \multicolumn{1}{c}{bind} & time \\ \hline
1 & 16.03 & 15.02 & 2.32 & \multicolumn{1}{c}{37.57} & 0.026 \\
5 & 16.04 & 15.02 & 2.35 & \multicolumn{1}{c}{101.17} & 0.027 \\
\textbf{10} & \textbf{16.10} & 15.02 & 2.40 & \multicolumn{1}{c}{179.53} & 0.028 \\
15 & \textbf{16.10} & 15.02 & 2.42 & \multicolumn{1}{c}{258.29} & 0.035 \\
20 & 16.07 & 15.02 & 2.45 & \multicolumn{1}{c}{337.99} & 0.036 \\ \hline
\end{tabular}
\label{ablation1_table}
\caption{
Quality and processing time with respect to the number of nearest vertices ($N_{closest}$). As $N_{closest}$ increases, the rendering quality initially improves and then declines, while the processing time for both generation and rendering increases. We selected $N_{closest} = 10$ as the optimal value.
}
\end{table}

\textbf{Number of Nearest Vertices ($N_{closest}$).   }
We conducted a study to evaluate the effect of the key binding process parameter, \(N_{closest}\), on both avatar generation quality and reconstruction time. \(N_{closest}\) determines the number of FLAME vertices that influence the deformation of a single splat. As shown in \autoref{ablation1_table}, the quality converged to its highest value when \(N_{closest}\) was set between 10 and 15. This suggests that using too few splats fails to adequately capture the expression variations of the FLAME model in the changes of the Gaussian splats, while allowing too many FLAME vertices to influence the deformation negatively impacts the representation of subtle expression changes.
Regarding processing time, increasing \(N_{closest}\) resulted in longer avatar generation and rendering times. This increase is attributed to the computational complexity of 
$O(N_{FLAME} \times N_{closest})$.
Through these experiments, we selected $N_{closest} = 10$ as it provided an optimal balance between rendering quality and processing time, yielding converged quantitative performance with a generation time of approximately 197 seconds while maintaining support for real-time rendering.

~\\
\textbf{Gaussian Scale Adaptation.   }
To validate the effectiveness of our proposed Gaussian scale adaptation method, we conduct an ablation study comparing our full model with a variant where this feature is disabled. In the ablated version, the scales of the Gaussian splats remain fixed to their initial values throughout the animation, without adapting to the local mesh density changes. As summarized in \autoref{tab:ablation_scale}, our scale adaptation method yields a significant improvement in reconstruction quality, measured by PSNR, with only a minor impact on rendering speed. This demonstrates that dynamically adjusting splat scales is crucial for accurately representing the stretching and compression of the facial surface during expression changes, thereby leading to higher-fidelity results.

\begin{table}[t]
\centering
\renewcommand{\arraystretch}{1.3}
\label{tab:ablation_scale}
\begin{tabular}{@{}l|cc@{}}
\toprule
Method & PSNR ($\uparrow$) & Speed (FPS) ($\uparrow$) \\
\midrule
w/o Scale Adaptation & 15.97 & \textbf{38.46} \\
\textbf{w/ Scale Adaptation (Ours)} & \textbf{16.10} & 35.71 \\
\bottomrule

\end{tabular}
\caption{Ablation study on the Gaussian scale adaptation. Our proposed method with scale adaptation shows a significant improvement in PSNR with a negligible drop in rendering speed.}
\vspace{-15pt}

\end{table}

~\\
\textbf{Loss Weighting in the Optimization Phase.   }
We also conducted an ablation study to evaluate how the relative weighting between Chamfer distance ($\lambda_c$) and Landmark loss ($\lambda_l$) affects avatar reconstruction quality. As shown in Table~\ref{tab:loss_ablation}, the combination of $\lambda_c = 0.01$ and $\lambda_l = 0.99$ achieved the highest PSNR value among all tested settings.
These results indicate that combining geometric alignment of the entire 3D head model with precise facial landmark alignment leads to more effective fitting. Through this ablation, we identified an optimal loss balance that improves overall quality.

~\\
\textbf{COO Formatting During the Binding Phase.   }
We evaluated the effect of adopting the COO format, which leverages the fact that the binding template \(\mathbf{B}\) contains sparse elements, by measuring the avatar generation time and rendering time. When the COO format was not used, the avatar generation time increased from 197 seconds to 248 seconds. Furthermore, the rendering time per frame increased dramatically from 0.028 seconds to approximately 21 seconds.
The COO format theoretically simplifies the computational complexity of our method from $O(N_{\text{FLAME}} \times N_{\text{splat}})$ to $O(N_{\text{FLAME}} \times N_{closest})$,
demonstrating its effectiveness. As a result of employing COO formatting, we not only reduced the avatar generation time but also enabled real-time rendering.

\begin{table}[t]
\centering
\renewcommand{\arraystretch}{1.3}
\label{tab:loss_ablation}
\begin{tabular}{cc|c}
\toprule
$\lambda_c$ (Chamfer) & $\lambda_l$ (Landmark) & PSNR $\uparrow$ \\
\midrule
0.00   & 1.00 & 16.01 \\
\textbf{0.01}   & \textbf{0.99}  & \textbf{16.10} \\
0.10  & 0.90  & 15.57 \\
0.50  & 0.50  & 15.58 \\
0.90  & 0.10  & 15.87 \\
0.99  & 0.01   & 15.67 \\
1.00 & 0.00   & 15.69 \\
\bottomrule
\end{tabular}

\caption{Ablation study on loss weighting for Chamfer and Landmark loss during the fitting process. The optimal combination weights were identified through this ablation.}
\vspace{-15pt}
\end{table}
\section{Conclusion}




In this paper, we propose S-Avatar, a one-shot framework for 3D head avatar reconstruction from a single image. Our method follows a three-phase pipeline, involving (1) generating the initial splats from a single image using a Gaussian generation module, (2) optimizing FLAME parameters to fit the FLAME head model on the initial splats, and (3) computing a binding template that adjusts the transformation of the Gaussian splats based on variations in the nearby FLAME vertices. This enables the generation of animatable, novel-view renderable 3D avatars from a single RGB image.
Through extensive evaluations including ablation study, we demonstrate the effectiveness and robustness of S-Avatar in terms of rendering quality for both new facial expressions and novel views. Furthermore, we validate our system design and capability for real-time rendering.
Unlike prior neural rendering methods that directly optimize representations from limited views, our approach first constructs a canonical head model and then controls expression and pose via binding. This decoupled design better preserves identity and structural consistency, especially under extreme viewpoint changes. Such simplification at the reconstruction stage enhances the accessibility of personal avatar generation, making it more practical for VR applications without the need for specialized equipment or prolonged capture sessions.


~\\ 
\textbf{Limitations and Future work.   }
Despite the advantages of simplified input and realistic facial expression rendering, our method has several limitations.
Firstly, our approach relies on the results of initial 3DGS generation process. If the initial splats are suboptimal, the final rendering quality is degraded regardless of subsequent processing steps.
To further enhance the robustness of initial splat generation, future work may incorporate fine-tuned modules derived from recent state-of-the-art methods~\cite{lyu2024facelift, shen2025high}. Thanks to the modularity and simplicity of our single-image pipeline, such modules can be seamlessly integrated into our framework.
Furthermore, we plan to implement an integrated system that leverages internal sensors of commercial Head Mounted Displays (HMDs) to control features such as FLAME expression parameters, thereby enabling the generated head avatar to be effectively utilized in VR/AR/MR applications.

~\\ 
\textbf{Ethical Considerations.   }
While our approach lowers the barrier to avatar creation by working from a single image, it also carries risks of misuse, including the fabrication of deceptive content. We prohibit use of our technology for generating false or non-consensual media and allow reproduction of our method solely for research purposes conducted with proper consent and ethical oversight.

\acknowledgments{
This research was supported by National Research Council of Science and Technology(NST) funded by the Ministry of Science and ICT(MSIT), Republic of Korea(Grant No. CRC 21011) and Institute of Information \& communications Technology Planning \& Evaluation (IITP) grant funded by the Korea government(MSIT) (No.2019-0-01270, WISE AR UI/UX Platform Development for Smartglasses). Also, we was supported by Institute of Information \& communications Technology Planning \& Evaluation (IITP) under the metaverse support program to nurture the best talents (IITP-2023-RS-2022-00156435) grant funded by the Korea government(MSIT).}

\bibliographystyle{abbrv-doi-narrow}

\bibliography{template}

\clearpage
\appendix
\appendix
\clearpage

\twocolumn[{%
\vspace*{0.6cm}
\begin{center}
    {\LARGE \textbf{S-Avatar : Diffusion-Guided Gaussian Head Avatars from a Single Image}}\\[10pt]
    {\Large Appendix.}
\end{center}
}]
\section{VR Rendering Results}

\begin{figure}[]
  \centering
  \includegraphics[width=0.9\columnwidth,keepaspectratio]{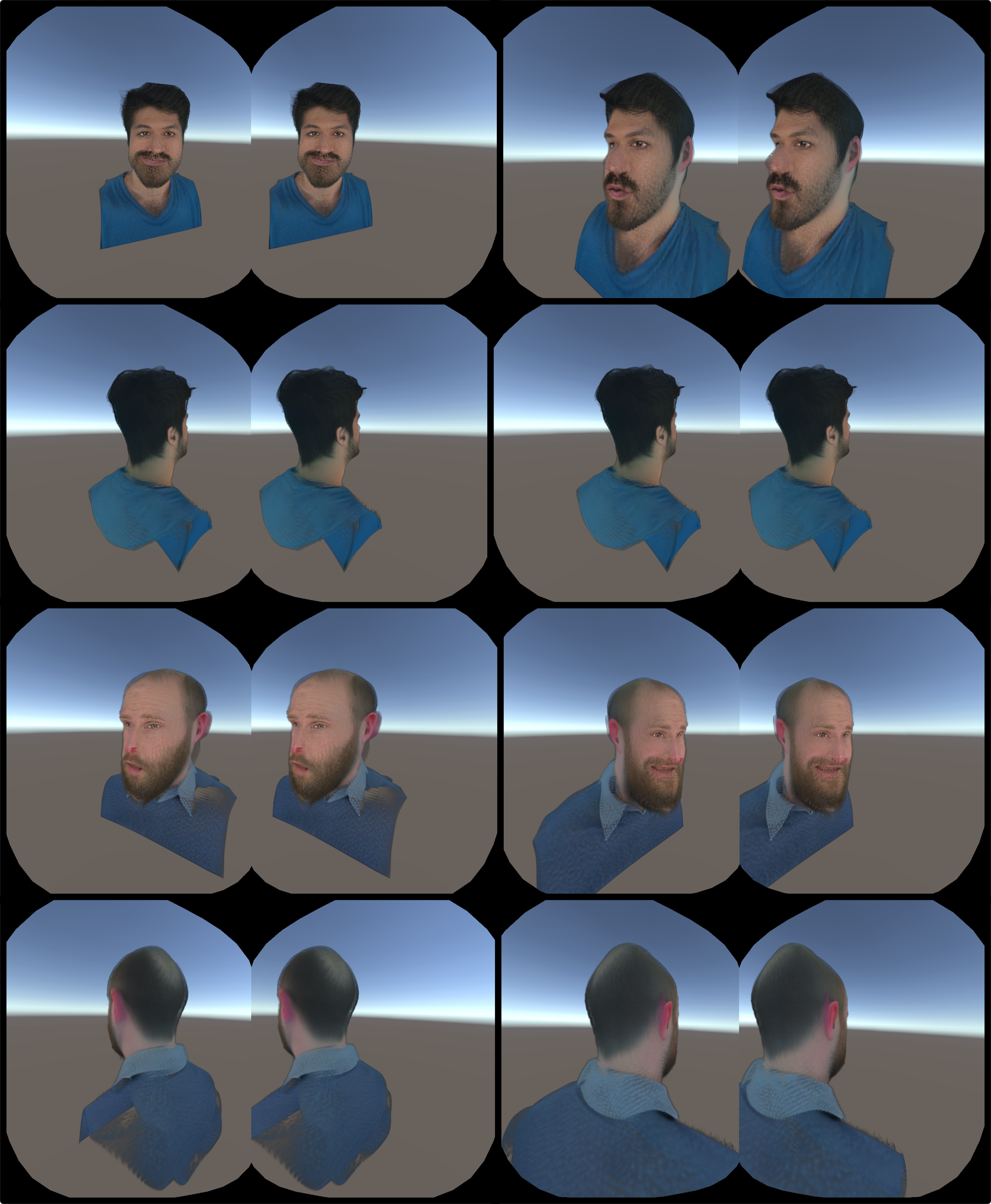}
  \caption{
Reconstructed avatars rendered on 
Meta Quest 3. Our method supports photo-realistic and stereoscopic rendering, enabling high-fidelity avatar representations suitable for VR/AR/MR applications.}
  \label{fig:vr_supp}
\end{figure}

\autoref{fig:vr_supp} shows the VR rendering results of a head avatar generated by S-Avatar. This VR rendering example demonstrates that our method can serve as an avatar in VR/AR/MR environments. The rendering results were processed with Unity\footnote{\label{Unity}\url{https://unity.com/}} and displayed stereoscopically through a commercial HMD, Meta Quest 3\footnote{\label{Meta Quest 3}\url{https://www.meta.com/quest/quest-3/}}.

\section{Interactive Viewer}

\begin{figure}[]
\centering
 \includegraphics[width=0.99\columnwidth]{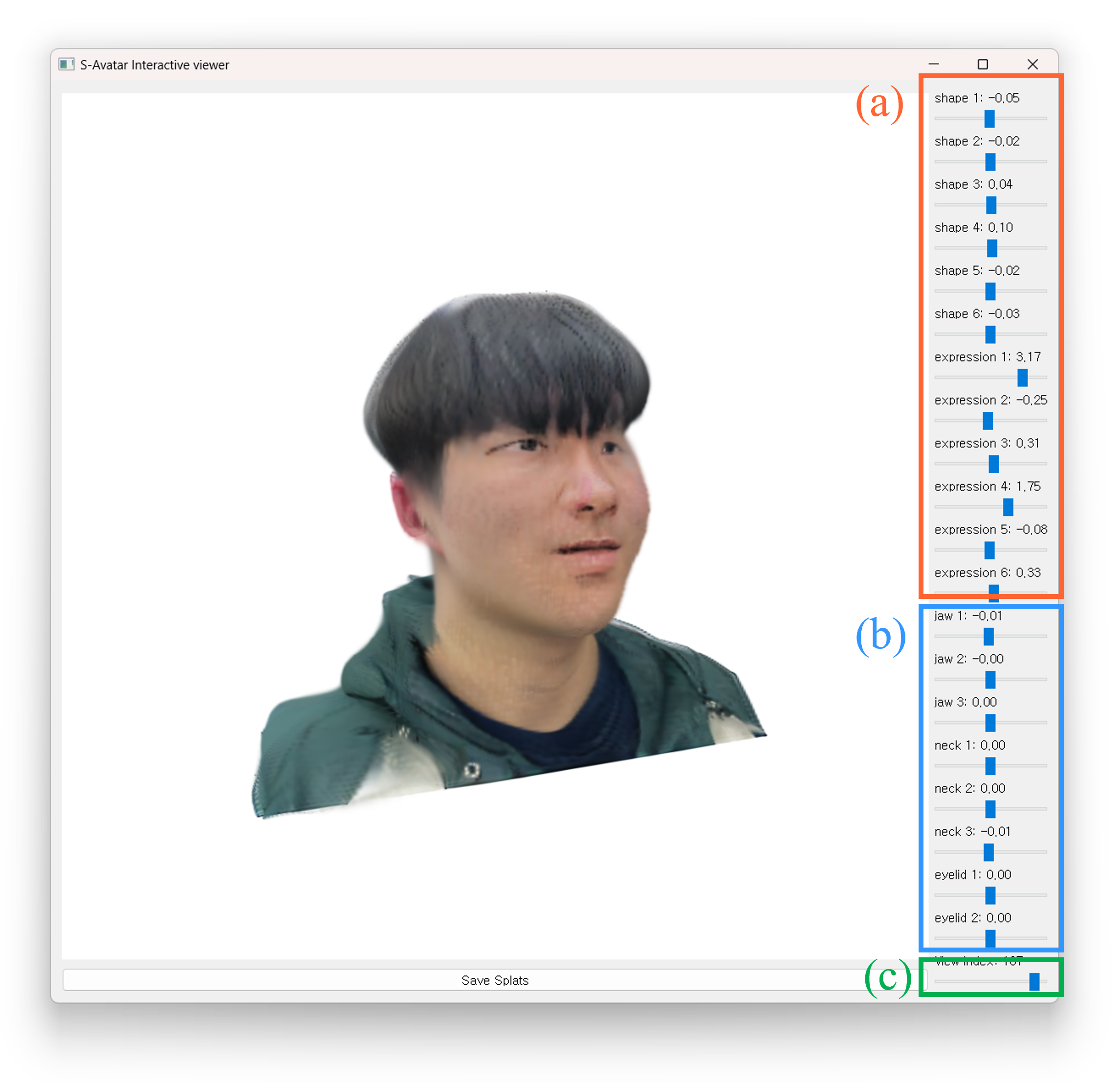}
 \caption{
    Interactive viewer for visualizing the output of S-Avatar. (a) FLAME parameter control via slider UI, allowing real-time adjustments of the avatar's shape and expression. (b) Joint rotation control for the jaw and neck. (c) Azimuth angle control at a fixed elevation for viewing the avatar from novel perspectives.
 }
 \label{fig:interactive}
\end{figure}

To facilitate intuitive visualization of rendering results, we developed an interactive viewer that supports real-time control and rendering of head avatars generated by our method, S-Avatar. \autoref{fig:interactive} presents the viewer, which allows users to manipulate the S-Avatar head model interactively. Using a slider-based UI, users can adjust FLAME parameters to observe changes in facial shape and expressions.
The viewer also supports control of jaw and neck joints through rotation values. Additionally, a final slider enables modification of the azimuth angle at a fixed elevation, allowing the exploration of novel viewpoints. The interactive viewer for S-Avatar is available at the project page\footnote{\label{project page}\url{https://github.com/hailsong/savatar}}.

\section{Real-time VR Demo}

\begin{figure}[]
\centering
 \includegraphics[width=0.99\columnwidth]{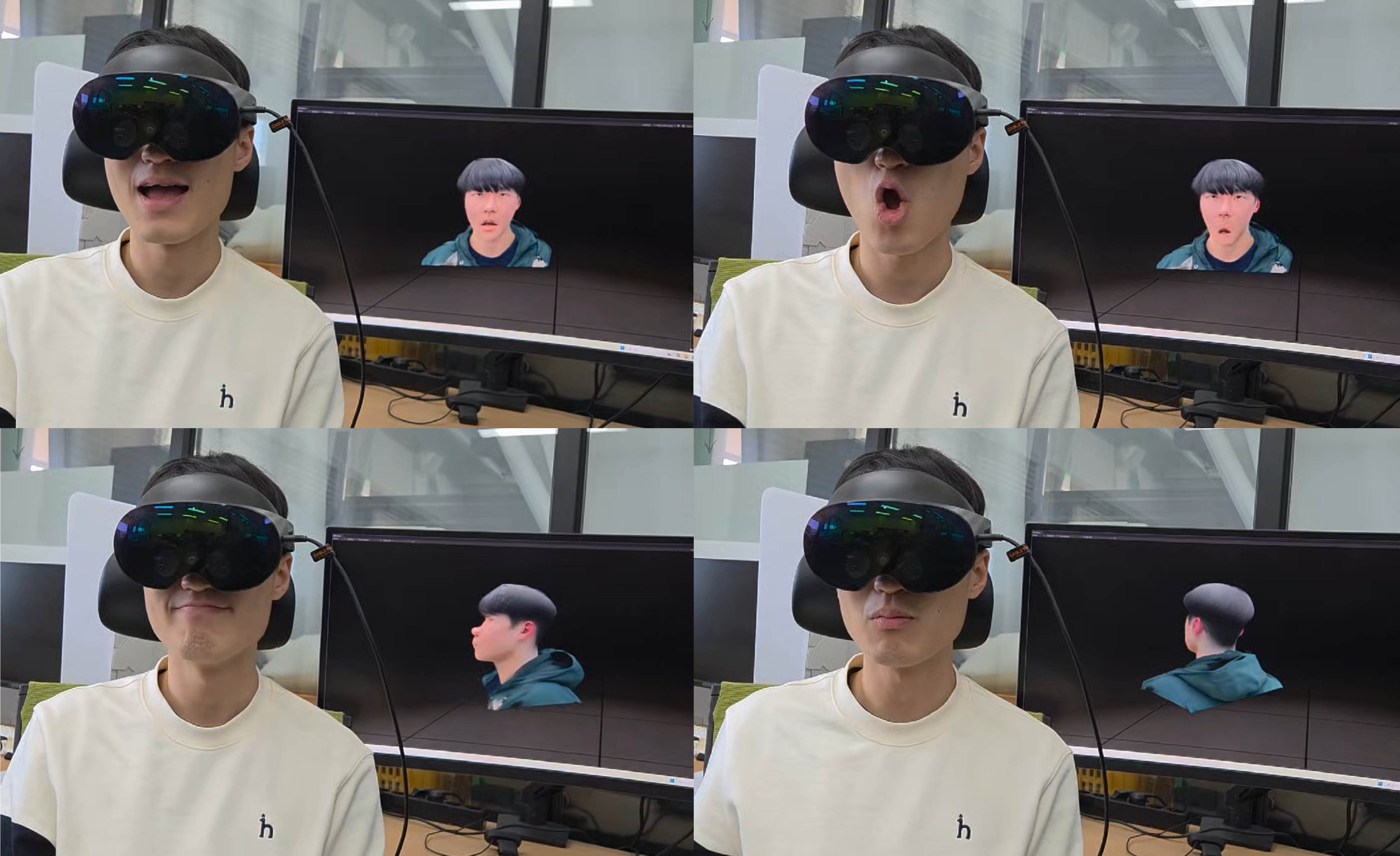}
 \caption{
    VR demo results using OFERA. The proposed method provides robust avatar reconstruction for the frontal view and unobserved viewpoints(side, back).
 }
 \label{fig:VR_Demo}
\end{figure}

To demonstrate the practical utility of our reconstruction approach, we present a VR demo application. We use OFERA\footnote{\label{project page}\url{https://ysshwan147.github.io/projects/ofera/}} to translate the internal sensor data from the HMD into FLAME expression parameters, enabling the reconstruction of a dynamic head avatar. Our method supports real time rendering and not only captures and reflects facial expressions from the frontal view in real time, but also facilitates robust avatar reconstruction for unobserved viewpoints(\autoref{fig:VR_Demo}), owing to the guidance of diffusion-based 3D modeling techniques.

\section{Text Guidance for Gaussian Generation Module}

To guide the Large multi-view Gaussian Model (LGM) process more robustly, we utilize carefully crafted text prompts as an additional modality of supervision. The full positive and negative prompts used to encourage photorealistic and high-fidelity head generation are as follows:

\begin{quote}
\texttt{Positive: A highly detailed, photorealistic image of a person's head. Captured with a professional camera using cinematic lighting, sharp focus, and realistic skin texture, rendered in 8K resolution with ultra-high detail.}
\end{quote}

\begin{quote}
\texttt{Negative: Noisy, blurry, or low-quality outputs; no cartoonish effects, text, or watermarks. Deformations, overexposure, unnatural CGI effects, and distorted facial features.}
\end{quote}

\begin{figure}[ht!]
\centering
\includegraphics[width=\columnwidth,keepaspectratio]{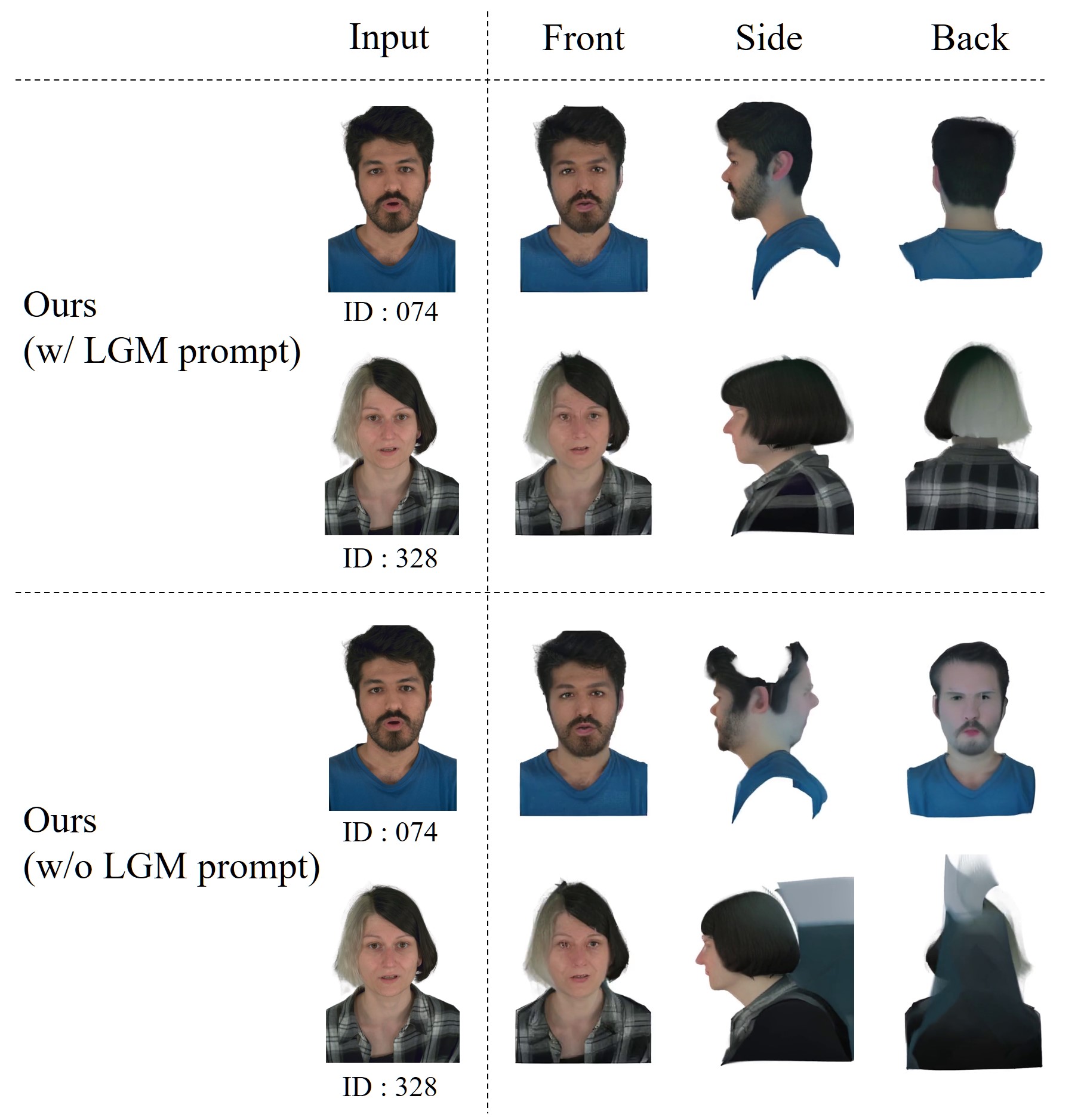}
\caption{
Rendering results of the initial splats $\mathbf{S}$ based on whether the text prompt input is provided for Gaussian generation module. It can be observed that providing a text prompt results in improved generation outcomes.
}
\label{fig:ablation_1}
\end{figure}

~\\
\textbf{Impact of Prompt on initial splats Generation.   }
To assess the impact of the optional text prompt input for Gaussian generation module, we observed the generation results of the initial splats with and without the use of positive and negative prompts. In the test dataset, we found that providing a text prompt generally resulted in a more robust generation of the 3D human head model. In particular, as shown in \autoref{fig:ablation_1}, subject ‘074’ exhibited a prominent failure case when no text prompt was provided—the loss of 3D consistency led to a Janus effect, where the face overlapped in areas that should have shown the back of the head. Additionally, for subject ‘328’, artifacts in the form of extra Gaussian splats appearing outside the head region were observed. These unintended Gaussian splats adversely affected facial landmark estimation and the fitting of the FLAME model, thereby hindering realistic avatar representation.

\section{Additional Qualitative Results}

\begin{figure*}[]
\centering
\includegraphics[width=\textwidth,keepaspectratio]{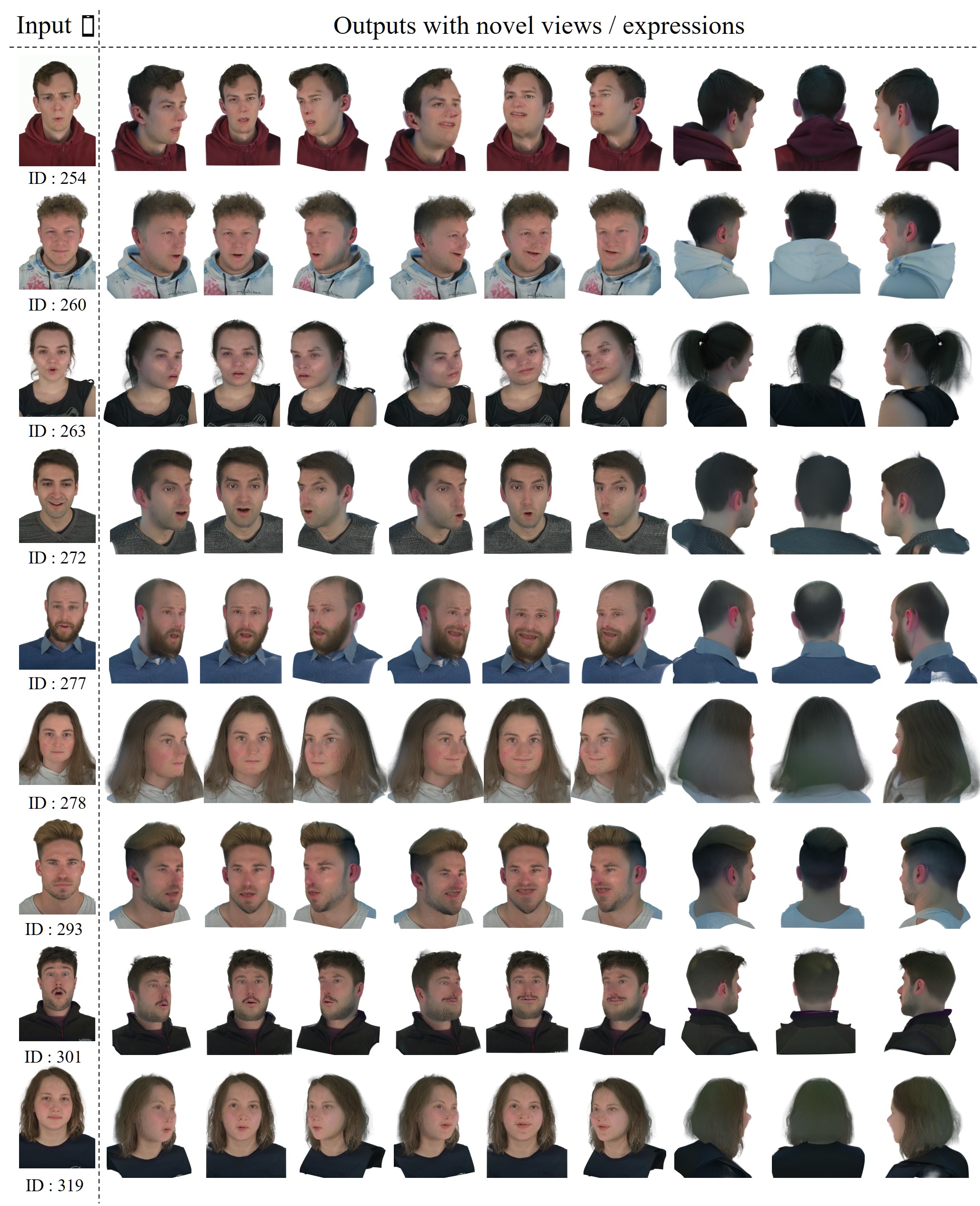}
\caption{
  Additional rendering results from our method, S-Avatar
}
\label{fig:qualitative_supp}
\end{figure*}

We present additional qualitative rendering results in \autoref{fig:qualitative_supp}. All avatars were generated using our method, S-Avatar, from a single frontal image input from the NeRSemble dataset. These results demonstrate that our method enables high-fidelity one-shot 3D head avatar generation.

\section*{Disclosure on Gen AI Usage}
In accordance with IEEE guidelines, the authors used Google Gemini for generating figure icons in the teaser, and ChatGPT for linguistic editing. All scientific content, experimental design, and data analysis remain the sole responsibility of the authors.

\end{document}